\documentclass[10pt,twocolumn,letterpaper]{article}

\usepackage{wacv}
\usepackage{times}
\usepackage{epsfig}
\usepackage{graphicx}
\usepackage[final]{pdfpages}
\usepackage{amsmath}
\usepackage{amssymb}
\usepackage{float}

\wacvfinalcopy % *** Uncomment this line for the final submission

\ifwacvfinal\fi
\begin{document}

%%%%%%%%% TITLE
\title{ Fusion of Methods Based on Minutiae, Ridges and Pores for Robust Fingerprint Recognition}

% Authors at the same institution
%\author{First Author \hspace{2cm} Second Author \\
% Authors at different institutions
\author{Lucas Alexandre Ramos \\
Unesp - Brazil\\
{\tt\small}
\and
Aparecido Nilceu Marana\\
Unesp - Brazil\\
{\tt\small }
}

\maketitle
\ifwacvfinal\thispagestyle{empty}\fi

%%%%%%%%% ABSTRACT
\begin{abstract}
  The use of physical and behavioral characteristics for human identification is known as biometrics. Among the many biometrics traits available, the fingerprint is the most widely used. The fingerprint identification is based on the impression patterns, as the pattern of ridges and minutiae, characteristics of first and second levels respectively. The current identification systems use these two levels of fingerprint features due to the low cost of the sensors. However, due the recent advances in sensor technology, it is possible to use third level features present within the ridges, such as the perspiration pores. Recent studies have shown that the use of third-level features can increase security and fraud protection in biometric systems, since they are difficult to reproduce. In addition, recent researches have also focused on multibiometrics recognition due to its many advantages. The goal of this work was to apply fusion techniques for fingerprint recognition in order to combine minutiae, ridges and pore-based methods and, thus, provide more robust biometrics recognition systems. We evaluated isotropic-based and adaptive-based automatic pore extraction methods and the fusion of pore-based method with the identification methods based on minutiae and ridges. The experiments were performed on the public database PolyU HRF and showed a reduction of approximately 16\% in the Equal Error Rate compared to the best results obtained by the methods individually.

\end{abstract}

%%%%%%%%% BODY TEXT
\section{Introduction}

Due to the increasing demands for safety and the increase in sensors capacity, a great number of studies with the third-level fingerprint characteristics have been carried out, as in \cite{angeloni}, where the fingerprint fragments recognition method based on ridges proposed by \cite{marana} was improved using the sweat pores detected on the ridges in the registration and comparison steps, through its fusion with the ridges recognition method.

Some fields such as forensics require systems with great precision and which are able to adapt to situations where the information on the fingerprints are incomplete and only fragments are available. According to \cite{jain}, the use of multibiometrics systems can significantly improve the accuracy of a system, making it safer and enabling better population coverage.

Adding pores to a fingerprint based biometrics systems can make it more secure, since the pores of fingerprints are difficult to reproduce, and studies indicate they can be used to distinguish fake silicon fingerprints from real ones, making the systems safer and more reliable  \cite{uludag}.

\begin{figure*}
\begin{center}
\includegraphics[width=0.9\linewidth]{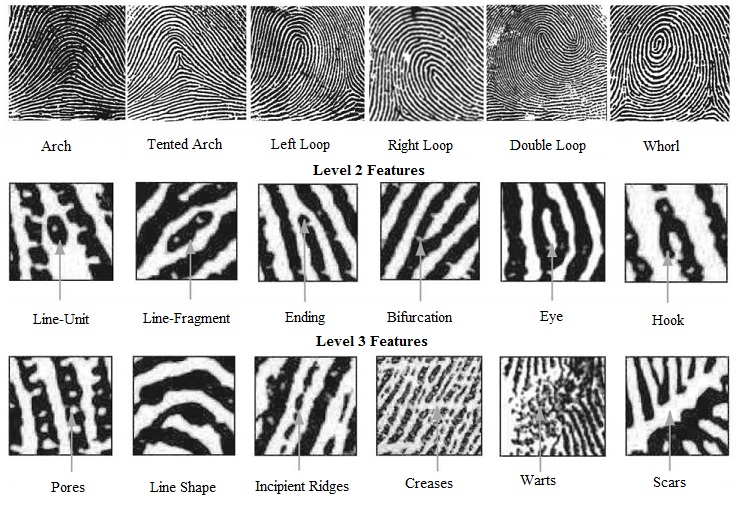}
\end{center}
   \caption{Fingerprint features at Level 1 (upper row), Level 2 (middle row), and Level 3 (lower row).\cite{jain2007}}
\label{feat}
\end{figure*}

According to \cite{jain11} the concept of biometrics is defined as the recognition of individuals based on their physical or behavioral characteristics, being the fingerprint one of the most commonly used biometric characteristics. The fingerprints presents three levels of characteristics: the patterns formed by macro details of the fingerprint, as the flow of the ridges (level one features), the minutiae formed by distinctive points observed in the flow of the ridges, as bifurcations and endings (level two features) and details extracted from ridges, such as the pores and scars (level three features) \cite{jain11}. Figure \ref{feat} shows the levels of fingerprint characteristics.

Currently only characteristics from level one and two are used on recognition systems \cite{jain2007}, \cite{zhao}. This is due the fact that the fingerprint images are generally obtained with resolutions up to 500 dpi, which is lower than the resolution published by NIST (National Institute of Standards and Technology, USA) in 2007 on its guidelines and standards for the third level features fingerprint recognition. NIST suggests the adoption of images with a resolution of at least 1000 dpi to allow the proper extraction of these characteristics \cite{oliveira}.

The minutiae-based systems are the most used ones specially because they employ the method adopted by forensic experts all over the world and the minutiae is accepted as proof of identity in virtually all countries \cite{maltoni}.

The goals of this paper are: to implement and evaluate fusion techniques for fingerprints recognition methods based on minutiae, ridges and pores and to perform the multibiometric recognition using partial fingerprint images.

%-------------------------------------------------------------------------
\section{Fingerprint Recognition Method Based on Minutiae}

This approach involves the use of a feature vector whose elements are the descriptors of the minutiae. Therefore, this approach’s methods are based on matching points algorithms that consists in finding a transformation (translation, rotation, and scale), such that the set of points of the reference image corresponds to the set of points of the query image.

The minutiae-based recognition method was implemented using the biometric systems technology called VeriFinger, which is distributed by a company called Neurotechnology.

VeriFinger uses a set of points (minutiae) for fingerprint recognition and provides a set of algorithms that increase performance and reliability. Some of their characteristics are, tolerance to deformation, rotation and translation and filters capable of eliminating noises and improving the quality of deteriorated images. Detailed information about the fingerprint recognition method based on minutiae is not provided by VeriFinger.

\subsection{Fingerprint Recognition Method Based on Ridges}

This fingerprint recognition method based on ridges proposed by Marana and Jain~\cite{marana}, uses the Hough Transform and the feature extraction stage consists of three steps. First the extraction and thinning of ridges based on the algorithm proposed by Jain~\cite{jain1997}, where the ridges are extracted after estimating the ridges orientation field and background segmentation. The second step is the straight line extraction, where the most significant straight lines from the fingerprint ridge pixels are extracted using the Hough Transform. Finally, the last step is the classification of the ridges according to their lines curvature. Figure 2 illustrates the steps for ridge extraction.

\begin{figure}[h]
\begin{center}
\
  \includegraphics[width=0.8\linewidth]{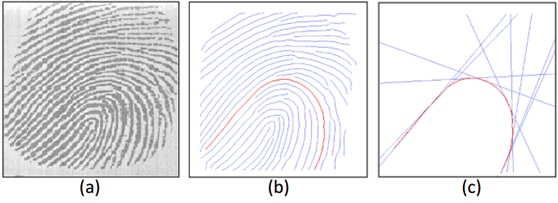}
\end{center}
   \caption{Ridge Extraction: (a) Fingerprint image; (b) Detected and thinned ridges from the fingerprint image; (c) Straight lines detected from a given  fingeprint ridge by using the Hough Transform Hough Transform \cite{marana}.}
\label{fig:onecol}
\end{figure}

The comparison consists of two steps, the fingerprint registration, where the parameters of geometric transformations (rotation, translation and scale) are calculated from the Hough Transform using the Hough space peaks, and the comparison stage where the query image is aligned to the reference image and the comparison score is calculated based on the number of matching ridges between these two image using a alignment matrix of $M \times N$ dimension where $M$ and $N$ are the number of ridges detected in query image and the reference image, respectively.

\section{Fingerprint Recognition Method Based on Ridges and Pores}

\begin{figure}[h]
\begin{center}
\
  \includegraphics[width=0.8\linewidth]{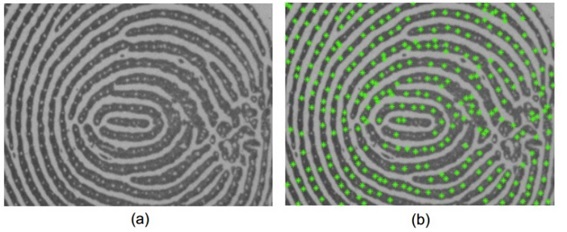}
\end{center}
   \caption{Pores extracted from a fingerprint fragment using the method based on adaptative filters proposed by Zhao~\cite{zhao}.}
\label{adap}

\end{figure}

\begin{figure}[h]
\begin{center}
\
  \includegraphics[width=0.8\linewidth]{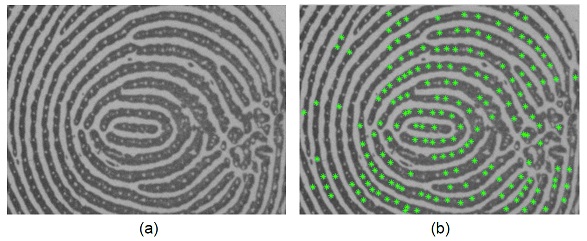}
\end{center}
   \caption{Pores extracted from a fingerprint fragment using the method based on isotropic filters proposed by Ray\cite{ray}.}
\label{isot}

\end{figure}

The recognition method based on ridges and pores proposed by \cite{angeloni}, aimed to extend the recognition method based on ridges \cite{marana}, adding the information from the pores of the fingerprint on the comparison step. The methods used for the pore extraction step are the following: the method based on adaptative filters proposed by Zhao~\cite{zhao}, and the method based on isotropic filters proposed by Ray\cite{ray}. Figure \ref{adap} shows the pores extracted from an image using the adaptative filter algorithm and Figure \ref{isot} shows the pores extracted using the isotropic filter algorithm.

In the strategy proposed by Angeloni~\cite{angeloni}, first the ridge-based method is executed and the transformation parameters obtained from the alignment stage are applied to the coordinates of the pores extracted from the query image, which acts as an additional heuristics in selecting the best alignment, then the pore comparison step is initiated. The scores obtained by the two methods are fused using a weighted sum, thus, obtaining the final score of the comparison. Figure 5 shows a diagram of the proposed strategy.

\begin{figure}[h]
\begin{center}
  \includegraphics[width=0.8\linewidth]{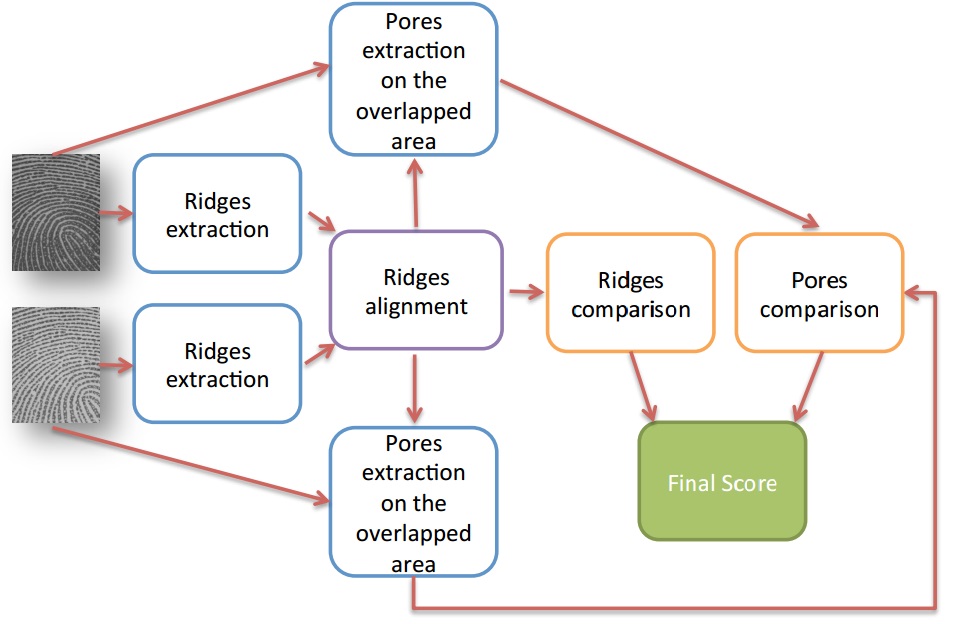}
\end{center}
   \caption{Diagram of the method proposed by Angeloni~\cite{angeloni}.}

\label{diagram}
\end{figure}

In the pore comparison stage a bounding box of 6, 8 and 10 pixels is used around each pore extracted due to fingerprint natural distortion. Thus, if after the alignment transformation a pore is inside the bounding box of the query image, then it would be considered a match.

%-------------------------------------------------------------------------
\section{The Fingerprint Database}

The PolyU HRF \cite{Polyu} is a public database provided by the Hong Kong Polytechnic University. The DB PolyU HRF I dataset consists of 1480 images, 10 images collected for 148 people in two separate sections. Images were captured using a high-resolution sensor, with resolution of 1200 dpi and dimension of 240x320 pixels.

\begin{figure}[t]
\begin{center}
  \includegraphics[width=0.8\linewidth]{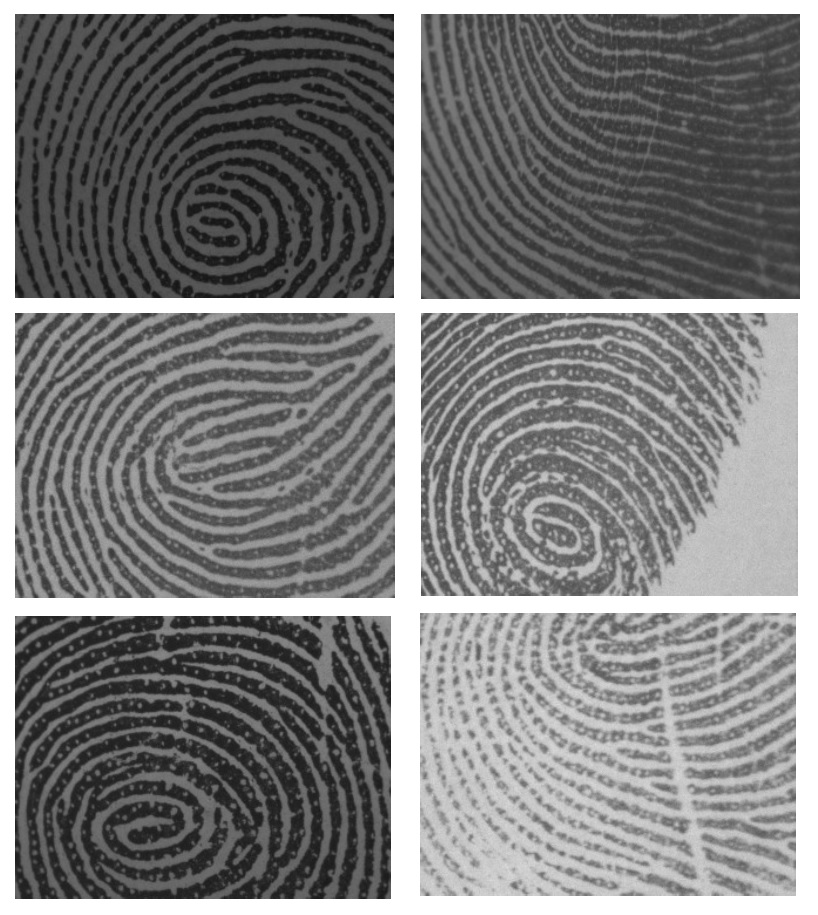}
\end{center}
   \caption{Examples of fingerprints fragments found in the PolyU HRF database.}
\label{pol}

\end{figure}

The PolyU HRF database is quite challenging because in addition to presenting common problems like non-linear distortion, rotation and translation, the area of comparison of the fingerprint fragments is very small. Figure \ref{pol} shows some fingerprints fragments found in this database.

The experiments were performed according to the protocol used by Zhao~\cite{zhao2009}, which consists of two types of comparisons, namely:

Genuine comparisons: The database is composed of five samples of fingerprints of each individual, obtained in two different sessions, in a total of 10 images per individual. For the genuine fingerprint comparisons, each fingerprint image of the second session was compared with all images of the same fingerprint of the first session, totaling 3700 genuine comparisons.

Impostor comparisons: For the impostor fingerprint comparison tests the first fingerprint of each finger from the second session was compared with the first image from first session of all the other fingerprints totaling 21756 impostor comparisons.

\section{Experimental Results}

In the experiments we analyzed the recognition accuracy using the following combinations: i) only minutiae; ii) only ridges; iii) only pores; iv) minutiae and ridges; v) ridges and pores; vi) minutiae and pores; and vii) minutiae, ridges and pores. For the results evaluation the Equal Error Rate (EER) was calculated for each experiment refered previously. All the methods were fused using a score-level fusion technique using weights for each score. The final weights were chosen on all possible combinations. Table \ref{tb1} shows the individual results obtained by each method before fusion. Table \ref{tb2} shows the results obtained for the fusion in pais (ridges and pores, ridges and minutiae and finally pores and minutiae).

\begin{table}[H]
\begin{center}
\begin{tabular}{|l|c|}
\hline
Method & EER \\
\hline\hline
Minutiae based & 25,08\% \\
Ridges based & 23,50\% \\
Pores Based (Isotropic) & 26,02\%\\
Pores Based (Adaptative) & 23,22\%\\

\hline
\end{tabular}
\end{center}
\label{tb1}
\caption{EER obtained for each method individually.}
\end{table}

\begin{table}[H]
\begin{center}
\begin{tabular}{|l|c|}
\hline
Fused Methods & EER \\
\hline\hline
Ridges and Pores Isotropic & 22,01\% \\
Ridges and Pores Adaptative & 22,31\% \\
Minutiae and Ridges & 9,35\% \\
Minutiae and Pores (Isotropic) & 10,45\%\\
Minutiae and Pores (Adaptative) & 9,08\%\\

\hline
\end{tabular}
\end{center}
\label{tb2}
\caption{Fusion of methods in pairs (minutiae, ridges and pores).}
\end{table}

Finally, the fusion of the three methods was performed. The weights used in the weighted sums were obtained using all possible combinations. Table \ref{tb3} shows the best results for the three methods fusion.

\begin{table}[H]
\begin{center}
\begin{tabular}{|l|c|}
\hline
Fused Methods & EER \\
\hline\hline
Minutiae, Ridges and Pores( Isotropic) & 8,57\%\\
Minutiae, Ridges and Pores (Adaptative) & 8,74\% \\

\hline
\end{tabular}
\end{center}
\label{tb3}
\caption{Fusion of methods based on Minutiae, Ridges and Pores}
\end{table}

It's important to stress that the best results were obtained with a higher weight assigned to the recognition method based on minutiae.

\section{Conclusion}

The recognition of fingerprint fragments proved to be quite challenging, even when using several methods, and it is of great importance in many areas, particularly forensics, where fingerprints fragments are common and their identification is essential.

The great accuracy improvement is due to the low amount of minutiae found in the fingerprints fragments, while the pores and ridges are abundant and may cause misclassification errors, minutiaes can be used as a confirmation whether the fingerprint if genuine or impostor, reducing therefore the false accpetance rate. The recognition method based on minutiae proved to be very accurate in identifing impostor fingerprint fragments, however it was not very efficient in identifying genuine ones, proving that commercial systems also have difficulty in recognizing fingerprint fragments and may benefit from the fusion of multiple fingerprint characteristics .

The recognition methods based on ridges and pores presented more balanced results, with both showing problems in identifying genuine and impostors fingerprint fragments, and are seldom used commercially, mainly because their study is recent, and its sensors are highly expensive.

The fusion of the methods proved to be quite effective, decreasing by more than 16\% the EER when minutiae, ridges and pores were fused making the system more reliable. The study of multibiometrics fusion proved to be very promising and relevant, since the fingerprint is the most widely used biometric characteristics used for identifing individuals.

%\includepdf[pages=-]{Reviews.pdf}

\section{Acknowledgement}
The authors would like to thank Fapesp for the financial support.

\end{document}